\PassOptionsToPackage{pdftex,dvipsnames}{xcolor}
\documentclass[letterpaper, 10 pt, conference]{ieeeconf}  

\IEEEoverridecommandlockouts                             
\usepackage{amsmath}
\usepackage{amssymb}
\let\labelindent\relax
\usepackage{enumitem}

\usepackage{amsmath,amsfonts}
\usepackage[font=footnotesize]{caption}
\usepackage{subcaption}
\usepackage{algorithm}
\usepackage{algpseudocode}
\usepackage{array}
\usepackage{textcomp}
\usepackage{url}
\usepackage{graphicx}
\usepackage{tabularray}
\usepackage{float}
\usepackage{bm}
\usepackage{mathtools}
\usepackage{stfloats}
\usepackage{mathrsfs}
\usepackage{multirow}
\usepackage{makecell}
\usepackage{vcell}
\usepackage{booktabs}
\usepackage{diagbox}
\usepackage{csquotes}
\usepackage{tabularx}
\usepackage{tikz}
\usepackage{lipsum}
\usepackage{url}
\usepackage{schemata}
\usepackage{verbatim}
\usepackage{pifont}

\usepackage{cite}
\makeatletter
\let\NAT@parse\undefined
\makeatother
\usepackage{hyperref}
\hypersetup{
colorlinks=true,
linkcolor=blue,
filecolor=magenta,
urlcolor=blue,
}
\hypersetup{breaklinks=true}
\begin{document}

\title{\LARGE \bf
Task Structure and Scalability in Multi-Robot Authoring
}

\author{
    Christian Marowski\\
    Indiana University Bloomington
    \and
    Byung-Cheol Min\\
    Indiana University Bloomington
}

\author{Christian Marowski, and Byung-Cheol Min
\thanks{Christian Marowski and Byung-Cheol Min are with the
SMART Lab, Luddy School of
Informatics, Computing, and Engineering, Indiana University
Bloomington, Bloomington, IN, USA.
{\tt\small \{cmarows@iu.edu,minb\}@iu.edu}}%
}

\maketitle
\begin{abstract}
Multi-robot authoring can reduce completion time by distributing drawing work, but the resulting speedup depends on task structure and the allocation-routing pipeline. We investigate these relationships by representing drawings as geometric stroke graphs and comparing component-level (single-token) allocation with shared-token (decomposed) allocation, which permits multiple robots to work within a connected component. A simulation study of 385 drawings and teams of one to eight robots shows diminishing speedup gains under single-token allocation and larger mean speedups under the shared-token pipeline, each normalized by its own single-robot completion time. Shared-token execution also exhibits lower idle-time shares at larger team sizes, although it does not reduce completion time for every drawing. Comparisons across structural attribute groups reveal associations between structural descriptors and differences in scaling behavior, with greater separation between several attribute groups at larger team sizes. These findings characterize empirical scaling in the evaluated pipelines and highlight the importance of considering within-component workload and graph structure alongside component count. Additional materials are available at \url{https://sites.google.com/view/multirobotauthoring/home}.
\end{abstract}

\vspace{-1pt}
\section{Introduction}
Multi-robot authoring uses teams of robots to reproduce user-specified geometry through drawing, printing, and related fabrication processes~\cite{lee2006multi,zhang2018large,santos2020interactive}. Additional robots create opportunities for parallel execution, but do not guarantee proportional reductions in completion time. Limits on parallel work are a familiar concern in parallel computing~\cite{amdahl1967validity}, while multi-robot execution also involves travel, workload imbalance, and interference~\cite{hazon2008redundancy}. For authoring, the central question is therefore: \emph{How do task structure and allocation strategy affect the benefit of increasing team size?}

\begin{figure}[t]
\centering
\includegraphics[width=0.90\columnwidth]{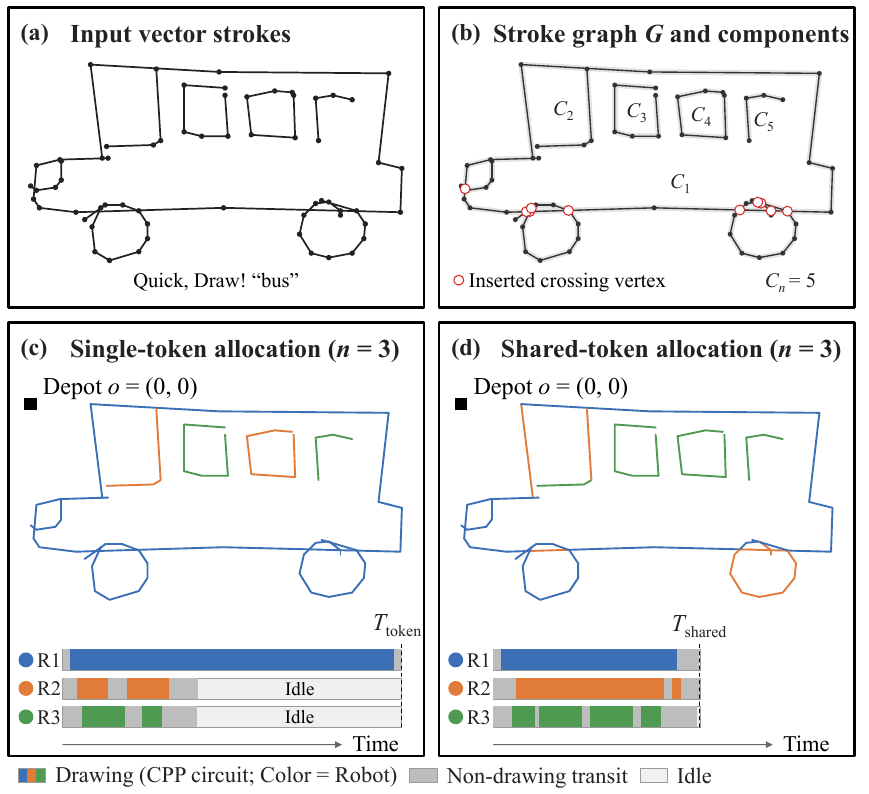}
\vspace{-3pt}
\caption{Task structure and allocation granularity on a Quick, Draw! drawing (``bus''). (a) Input vector strokes. (b) Stroke graph with inserted crossing vertices (red) and connected components. (c) Single-token allocation assigns each component to one robot. (d)~Shared-token allocation splits work inside a component across robots. Bars show schematic execution timelines.}
\vspace{-5pt}
\label{fig_1}
\end{figure}

A geometric stroke graph makes this question explicit (Fig.~\ref{fig_1}). Edges represent drawable segments, and vertices represent stroke waypoints and detected intersections. Under component-level (single-token) allocation, each connected component is assigned as an indivisible task to one robot. Under shared-token allocation, work within a component can be distributed among robots. This distinction changes the available concurrency, but the resulting completion time also depends on route construction, non-drawing transit, and local robot interactions.

We evaluate 385 user-generated drawings with one, two, three, four, six, and eight robots in a shared simulator. We report completion time, speedup relative to each pipeline's own single-robot baseline, and robot-time composition. The shared-token pipeline exhibits larger mean speedups and lower idle-time shares at larger team sizes, but is not faster for every drawing. Comparisons across structural attribute groups describe variation in scaling differences. Because routing procedures also differ, the study characterizes complete allocation-routing pipelines rather than isolating the causal effect of allocation granularity. Our focus is to characterize how the relative performance of two implemented allocation-routing pipelines varies across such inputs, rather than to introduce a new printing platform or claim that decomposition alone determines performance.

The key contributions of this work are as follows:
\begin{itemize}[leftmargin=*]

\item An empirical scaling comparison of component-level and shared-token allocation-routing pipelines across 385 stroke graphs and six robot-team sizes.
\item A paired completion-time analysis and robot-time accounting that characterize performance differences across the two pipelines.

\item A descriptive analysis of associations between six structural descriptors and differences in the pipelines' speedup relative to their respective single-robot baselines.

\end{itemize}

\vspace{-5pt}
\section{Related Work}
\vspace{-3pt}

Mobile printing, collaborative deposition, and coverage-based painting demonstrate how robot teams can reproduce geometry in different physical media~\cite{lee2006multi,zhang2018large,santos2020interactive}. YouWasps allocates deposition trajectories among mobile robots~\cite{sustarevas2019youwasps}; SPRINTER and subsequent algorithmic work investigate swarm-printing hardware and decomposition into rasterized geodesic cells~\cite{karpe2019sprinter,karpe2021optimized}. Online cooperative printing also combines printing-task allocation with execution by mobile robots~\cite{alhijaily2023online}. Recent cooperative additive-manufacturing studies more directly address partitioning and scheduling. Poudel et al. formulate resource-constrained scheduling for cooperative printing~\cite{poudel2021resource}. NoodlePrint uses cellular segmentation with helically interlocked tiles~\cite{ebert2025noodleprint}. SafeZone* develops a graph-based cooperative printing framework that accounts for geometric and topological structure~\cite{stone2025safezone}, while Traveling cellsman couples partition and cluster parameters for multi-robot printing~\cite{ebert2025cellsman}. Heterogeneous swarm manufacturing further integrates task division, allocation, scheduling, and path planning across different manufacturing capabilities~\cite{wu2026heterogeneous}. These works make geometry-aware decomposition and coordination established concerns. Our study instead examines how predefined stroke-graph descriptors relate to the scaling of two allocation-routing pipelines across a heterogeneous two-dimensional drawing corpus.

Task-allocation taxonomies organize problems by task and robot requirements and by dependencies among assignments~\cite{gerkey2004formal,korsah2013taxonomy}. Market-based coordination provides a complementary allocation perspective~\cite{dias2006market}; work on complex tasks already considers decomposition and allocation together~\cite{zlot2006complex}. These perspectives motivate distinguishing how allocatable work is defined from how that work is assigned. Coverage planning provides related examples of dividing work among robots~\cite{galceran2013coverage}. Studies of redundancy and efficiency examine the costs of multi-robot coverage~\cite{hazon2008redundancy}, and DARP partitions a coverage area among robots~\cite{kapoutsis2017darp}. More recent grid-coverage work explicitly incorporates path deconfliction~\cite{tang2025coverage}. These methods address area or grid coverage, whereas stroke authoring requires traversal of specified drawing geometry and distinguishes drawing from non-drawing movement. Our execution layer uses local yielding without joint path replanning, so we do not equate nominal route balance with conflict-free completion time.

\begin{figure*}[t]
\centering
\includegraphics[width=0.97\textwidth]{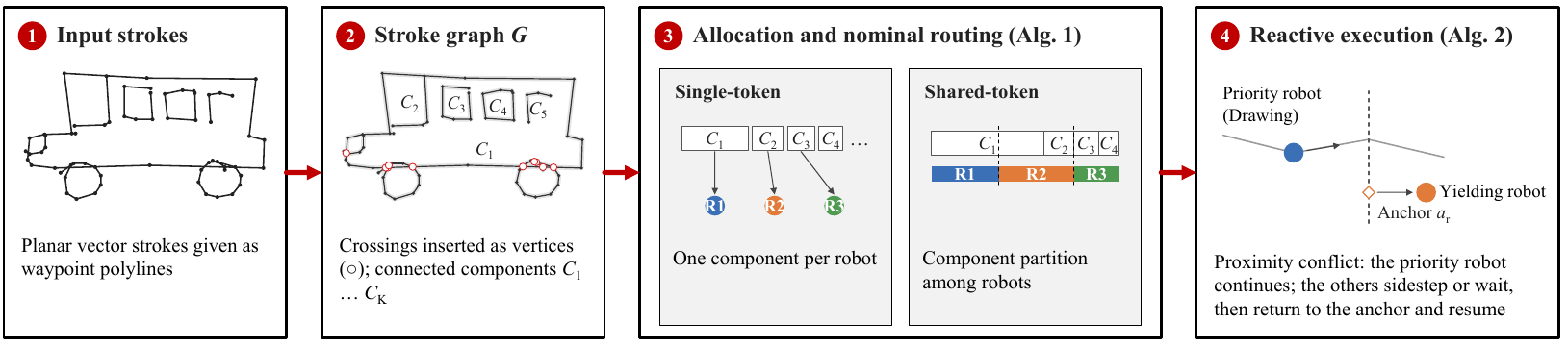}
\vspace{-5pt}
\caption{System overview. Vector strokes form a stroke graph. Single-token allocation treats each component as indivisible, whereas shared-token allocation permits work within a component to be distributed among robots (Algorithm~\ref{alg:authoring}). The reactive layer executes the routes using priority assignment and anchor-based yielding (Algorithm~\ref{alg:yield}).}
\vspace{-20pt}
\label{fig:overview}
\end{figure*}

The Chinese postman problem provides a basis for covering required edges with economical repeated traversal~\cite{edmonds1973matching}. Multi-route routing and tour-splitting methods provide related foundations for distributing traversal work~\cite{frederickson1978approximation}. Multi-robot line coverage is especially relevant because it distinguishes servicing required segments from deadheading and considers routing constraints and robot experiments~\cite{agarwal2024line}. Our shared-token planner constructs component CPP circuits, forms a fixed master traversal, and partitions that traversal. Exact partitioning of this fixed order does not imply globally optimal multi-postman routing. The paper's contribution is an empirical characterization of the resulting allocation-routing pipelines, using paired completion times, method-specific speedups, and robot-time composition. We distinguish direct completion-time gains from self-normalized scaling when relating performance to input structure.

\vspace{-5pt}
\section{Methodology}
\subsection{Input, Task Representation, and Scope}
\label{sec:representation}

Figure~\ref{fig:overview} gives an overview of the pipeline described in this section. The method takes a finite set of planar vector strokes, or an equivalent geometric stroke graph, as input. A vector stroke is a sequence of coordinates describing a polyline. For the experimental corpus, consecutive QuickDraw waypoints define vector segments. Geometric crossings and self-crossings are inserted as vertices using Shapely's \texttt{unary\_union}, while original waypoints are retained as graph vertices. Exactly overlapping linework is merged by the geometric union; no coordinate snapping or rounding is applied. The resulting saved graph is used directly by the simulator without further intersection processing. Conversion from raster images, semantic interpretation of handwriting, and hardware-specific control are outside the input representation used here. Let
\vspace{-5pt}
\begin{equation}
\vspace{-5pt}
G=(V,E), \qquad \mathcal{C}(G)=\{C_1,\ldots,C_K\}
\label{eq:graph}
\end{equation}
denote the graph and its connected components. Here $K$ denotes connected-component count, and $n$ denotes robot count. Every required edge $e\in E$ stores a geometric path $\gamma_e$ and length $\ell_e$. Intersection processing and the treatment of coincident geometry affect the topology of $G$ and, consequently, the component-level tasks available for allocation.

Let $P_r$ denote the route assigned to robot $r$, including drawable and non-drawing portions, and let $E_{\mathrm{draw}}(P_r)$ be the required edges reproduced in the drawing state. Completion of the authoring work requires
\vspace{-5pt}
\begin{equation}
\vspace{-5pt}
\bigcup_{r=1}^{n}E_{\mathrm{draw}}(P_r)=E.
\label{eq:coverage}
\end{equation}
This is a coverage condition, not an assertion that every geometric edge is traversed exactly once. Chinese postman problem (CPP) augmentation may require repeated traversal of graph edges, while movement between disconnected route portions and depot access are performed without drawing. The planner outputs routes and, for token allocation, subsequent task assignments during execution.

\subsection{Structural Descriptors}
\label{sec:descriptors}

We characterize workload, topology, and spatial structure before allocating the drawing. For component $C_c$, define
\vspace{-10pt}
\begin{equation}
L_c=\sum_{e\in E(C_c)}\ell_e,\quad
L=\sum_{e\in E}\ell_e,\quad
C_L=\frac{\max_c L_c}{L}.
\label{eq:workload}
\end{equation}
The workload descriptors are component count $K$, total required stroke length $L$, and component concentration $C_L$. A high $C_L$ identifies drawings whose nominal work is concentrated within one component. Cyclicity and branching are
\vspace{-5pt}
\begin{equation}
\vspace{-5pt}
\mu=|E|-|V|+K,\qquad
B=\sum_{v\in V}\max\{0,\deg(v)-2\},
\label{eq:topology}
\end{equation}
where $\mu$ is the cycle rank, and $B$ sums the vertex degrees in excess of two.

Spatial dispersion is computed using the unique original QuickDraw waypoints rather than intersection-only vertices introduced during graph construction. For waypoint coordinates $\{\mathbf p_i\}_{i=1}^{N}$, let $\bar{\mathbf p}=N^{-1}\sum_i\mathbf p_i$. Spatial dispersion and stroke density are
\vspace{-5pt}
\begin{equation}
\vspace{-5pt}
D=\frac{1}{N}\sum_{i=1}^{N}\|\mathbf p_i-\bar{\mathbf p}\|_2^2,
\qquad \rho_s=\frac{L}{A_{\mathrm{bbox}}},
\label{eq:spatial}
\end{equation}
where $A_{\mathrm{bbox}}$ is the area of the axis-aligned bounding box of the stored stroke geometry. Duplicate waypoint coordinates are counted once in $D$. The ratios require nonzero denominators; degenerate cases are retained as undefined rather than assigned an artificial value.

\subsection{Allocation and Nominal Routing}
\label{sec:allocation}

\subsubsection{Token allocation}

Each connected component is assigned to exactly one robot. With $x_{cr}\in\{0,1\}$ recording whether robot $r$ receives $C_c$, this restriction is
\vspace{-5pt}
\begin{equation}
\vspace{-5pt}
\sum_{r=1}^{n}x_{cr}=1,\qquad c=1,\ldots,K.
\label{eq:token}
\end{equation}
Eq.~\eqref{eq:token} expresses the assignment rule; it does not imply that the implementation solves an integer program.

The dispatcher maintains an ordered queue of unassigned components using their deterministic order in the saved graph. Initially, successive components are assigned to successive robots. A robot that completes its component immediately receives the next unassigned component, if one remains. For each assigned component, the robot selects the component vertex nearest its current position, moves to that vertex without drawing, and then follows an undirected CPP circuit covering that component. A robot moves directly from the end of one component to the next assigned component without first returning to the depot. After completing its work with no further task available, it returns to the designated depot. Robots that receive no work remain idle. Component order is not optimized according to component workload or estimated completion time.

\subsubsection{Shared-token (decomposed) allocation}

Work within a connected component can instead be distributed among robots. The implemented Shared-token planner follows a route-first, split-second formulation related to tour-splitting approaches for the min-max multi-postman problem \cite{frederickson1978approximation}. An undirected CPP circuit is first constructed independently for each connected component using shortest-path augmentation of odd-degree vertices and minimum-weight matching \cite{edmonds1973matching}. Component circuits are then concatenated into a fixed master traversal using a greedy nearest-component rule: from the current route endpoint, the next component is the unvisited component with the nearest graph vertex, which becomes the source of that component's CPP circuit.

Let $\mathcal{P}(G,n)$ denote the contiguous partitions of this fixed master traversal among $n$ robots, or fewer when the traversal contains fewer edge traversals than robots, and let $L_r(P_r)$ denote the nominal length of robot $r$'s resulting route, including drawing traversal, direct transit between disconnected portions, and travel from and back to the common depot. The implemented objective is
\vspace{-4pt}
\begin{equation}
\vspace{-4pt}
\min_{(P_1,\ldots,P_n)\in\mathcal{P}(G,n)}
\max_{1\le r\le n} L_r(P_r).
\label{eq:minmax}
\end{equation}
The partition is solved exactly by dynamic programming for the fixed master traversal, but this is not a claim of globally optimal min-max $k$-CPP routing because the master traversal and component order are fixed before partitioning. Partition boundaries may occur between graph-edge traversals inside the same connected component, allowing multiple robots to share a token, but an individual stored edge is not subdivided internally.

The component-indivisibility rule in~\eqref{eq:token} is therefore not imposed. Balancing nominal route cost also does not guarantee minimum executed completion time because local avoidance and waiting subsequently alter execution. Because Token allocation uses a dynamic component queue while Shared-token allocation partitions a fixed master route, their comparison is a pipeline-level comparison rather than an isolated change in decomposition granularity.

Algorithm~\ref{alg:authoring} summarizes the overall allocation and execution workflow for both strategies.

\begin{algorithm}[t]
\caption{Allocation and execution workflow}
\label{alg:authoring}
\begin{algorithmic}[1]
\Require Graph $G$, robot count $n$, mode $d$
\State Initialize robots at the common depot.
\If{$d=\mathrm{token}$}
  \State Queue components and assign them to available robots.
\Else
  \State Build component CPP circuits and concatenate the master traversal.
  \State Partition the traversal by the min-max objective in~\eqref{eq:minmax}.
\EndIf
\While{drawing or return is incomplete}
  \If{$d=\mathrm{token}$ and work remains}
    \State Assign the next component to each available robot.
  \EndIf
  \State Execute routes with reactive yielding and update coverage.
  \State Return robots with no remaining work to the depot.
\EndWhile
\end{algorithmic}
\end{algorithm}

\subsection{Reactive Execution and Interface Assumptions}
\label{sec:reactive}

The execution layer distinguishes drawing along required geometry from non-drawing transit. Let $\mathbf q_r(t)$ be the tracked reference position of robot $r$, and $\mathbf o$ the prescribed depot. Robots are modeled as homogeneous discs moving at fixed speed in a fixed-timestep simulation. A local proximity rule identifies conflicts based on current center distance. The trigger distance is the larger of the configured proximity radius and a one-step closure bound that accounts for possible relative closing motion during a timestep.

Pairwise proximity relationships are combined into connected conflict components, allowing interactions involving more than two robots to be handled jointly. Within each conflict component, one robot receives priority. A robot currently drawing receives priority over one performing non-drawing movement. If modes match, priority is given to the robot with the least remaining distance in its current active segment; robot ID provides a deterministic final tie-break. The same priority is retained while the local conflict persists.

Yielding robots preserve the point from which their planned traversal will resume by recording an anchor $\mathbf a_r=\mathbf q_r(t_{\mathrm{yield}}).$

The reactive protocol attempts to move a yielding robot perpendicular to the priority robot's current motion. When a sidestep direction is first selected, both perpendicular directions are evaluated using one-step predicted positions of the other active robots, and the feasible direction with greater minimum clearance is selected. If neither direction can be taken without predicted overlap, the robot waits instead. Once outside the conflict region, it continues waiting until the priority robot has passed and is no longer approaching its anchor. It then returns without drawing to the anchor and resumes its original route. Recovery motion remains subject to the same conflict procedure. Algorithm~\ref{alg:yield} summarizes this anchor-based yielding protocol for a single conflict component.

\begin{algorithm}[!htb]
\caption{Anchor-based yielding protocol}
\label{alg:yield}
\begin{algorithmic}[1]
\Require Conflict component $\mathcal S$
\State Select priority robot $p$ using DRAW/MOVE state,
\Statex \hspace{\algorithmicindent}remaining active-segment distance, and robot ID.
\For{each yielding robot $r\in\mathcal S\setminus\{p\}$}
  \State Record $\mathbf a_r\gets\mathbf q_r(t_{\mathrm{yield}})$ if not already anchored.
  \State Evaluate both perpendicular sidestep directions.
  \If{a sidestep without predicted overlap exists}
    \State Move using the direction with greater clearance.
  \Else
    \State Wait.
  \EndIf
  \State Wait for the priority robot to pass and conflict to clear.
  \State Return to $\mathbf a_r$ without drawing.
  \State Resume the original assigned traversal.
\EndFor
\end{algorithmic}
\end{algorithm}

Robots initially co-located at the shared depot are handled through a staging region in which departure is serialized by waiting rather than sidestepping. Finished robots at the depot are removed from subsequent active-obstacle checks.

The avoidance layer leaves high-level routes unchanged and performs no graph search, future-stroke inspection, multi-agent path finding, or route replanning. Its movement and waiting costs are therefore distinct from nominal route cost. The formulation does not establish formal collision avoidance, deadlock freedom, or minimum avoidance overhead. Physical overlaps are recorded if they occur. If an interaction remains unresolved, execution continues under the same local procedure until progress resumes or the simulation reaches its maximum duration, at which point the run is recorded as timed out. Reproducing this interface on a physical robot requires a compatible motion controller, drawing-state control, and localization; the dataset-independent description is not a claim of unrestricted hardware generality.

\vspace{-5pt}
\section{Experiments}
\vspace{-2pt}
\label{sec:experiments}

\subsection{Research Questions and Compared Pipelines}
We evaluate the following research questions:
\begin{enumerate}
    \item[\textbf{RQ1}] How does each allocation-routing pipeline's speedup change with robot count $n$?
    \item[\textbf{RQ2}] For which drawings and robot counts does the shared-token pipeline achieve a lower completion time than the single-token pipeline?
    \item[\textbf{RQ3}] Which stroke-graph characteristics are associated with differences in scaling behavior between the two pipelines?
\end{enumerate}

In the figures, \emph{Token} refers to single-token/component-level allocation and \emph{Decomposed} refers to shared-token/subgraph-level allocation. Both pipelines use the same drawing corpus and simulator. Their routing policies are not assumed identical, so RQ2 concerns a pipeline-level comparison rather than an isolated granularity intervention.

\subsection{Drawing Corpus and Preprocessing}
\label{sec:corpus}

Google Quick, Draw! supplies the user-generated vector drawings used for evaluation \cite{ha2017neural}. The selected dataset representation provides Ramer-Douglas-Peucker-simplified stroke waypoints. Following Section~\ref{sec:representation}, detected intersections are inserted as graph vertices and intersecting segments are split. Geometry is preserved relative to the selected representation; this statement does not claim preservation of raw, unsimplified human trajectories. NetworkX represents the graphs, and Shapely performs geometric intersection detection and splitting.

The evaluation uses 385 drawings selected using coverage-constrained random sampling to span variation in component count and other graph characteristics. The original corpus description reports at least one drawing from each object category. This corpus defines the empirical scope of the present results rather than establishing coverage of all vector-authoring inputs.

\subsection{Simulator and Experimental Configurations}
\label{sec:simulator}

Experiments use a fixed-timestep two-dimensional simulator implemented in Python, with NumPy for numerical computation and Matplotlib for visualization. Robots are homogeneous discs with fixed size and speed. Kinematic waypoint-based displacement updates omit low-level physical dynamics. The center of each disc is its drawing point. Travel between drawable regions and to or from the depot is executed without drawing.

The specified common depot is $\mathbf o=(0,0)$. Although robots are initialized at this coordinate, the depot is treated as a staging region rather than a physical configuration in which finite-size robots overlap. Departures are serialized. After completing their assigned route, robots return to the depot and are removed from subsequent collision detection. Time spent waiting for staged release is counted as avoidance time, while travel to and from the depot is counted as transit. A run terminates only after all required drawing is complete and all dispatched robots have returned to the depot.

Each graph is evaluated with both pipelines for $\mathcal N=\{1,2,3,4,6,8\}.$ This specifies $385\times6\times2=4{,}620$ graph-count-pipeline configurations before any repeated trials. Six robot-count settings are not six repetitions of an identical condition. Across all 4,620 runs, no simulations timed out. Geometric robot overlap occurred in 1,428 runs (30.9\%), with incidents increasing at larger team sizes. The reactive protocol therefore serves as a local conflict resolution heuristic rather than a guarantee of collision-free execution. Table~\ref{tab:settings} summarizes the simulator settings and experimental configurations used throughout the evaluation.

\begin{table}[h]
\centering
\caption{Simulator settings and experimental configurations.}
\vspace{-5pt}
\label{tab:settings}
\footnotesize
\begin{tabular}{p{0.42\columnwidth}p{0.48\columnwidth}}
\toprule
Setting & Value/status \\
\midrule
Drawing corpus & 385 graphs \\
Robot counts & $1,2,3,4,6,8$ \\
Robot model & Homogeneous discs; center draws \\
Depot & $(0,0)$ \\
Timestep, speed, disc radius & $0.02$s; $80$ units/s; $2.25$ units \\
Conflict handling & Reactive priority/yield protocol \\
Coordinate scale & Native QuickDraw coordinates \\
Repetitions & One deterministic run per config. \\
Timeout & $300$s \\
\bottomrule
\end{tabular}
\vspace{-5pt}
\end{table}
\subsection{Performance Measures and Aggregation}
\label{sec:evaluation-metrics}

System performance is primarily measured by run completion time, defined as the elapsed time required to finish all assigned drawing work and all depot returns. Separately, robot-time utilization is summarized by accumulating the time that individual robots spend drawing, making non-drawing movements, executing the reactive avoidance protocol, and idling. This accounting describes the use of robot resources rather than an additive decomposition of elapsed completion time. Drawing fidelity is not evaluated because the simulator constrains drawing motion to the predefined stroke geometry.

Let $T_m(G,n)$ denote the completion time for drawing $G$ using $n$ robots under pipeline $m\in\{\mathrm{token},\mathrm{shared}\}$. Scaling within each pipeline is measured by
\vspace{-5pt}
\begin{equation}
\vspace{-5pt}
S_m(G,n)=\frac{T_m(G,1)}{T_m(G,n)}.
\label{eq:speedup}
\end{equation}
Each pipeline is therefore normalized by its own single-robot completion time for the same drawing. At each robot count, the scaling curve reports the arithmetic mean of these per-drawing speedups. This is a mean of ratios, not a ratio of mean completion times.

The paired relative completion-time difference between the pipelines is
\vspace{-5pt}
\begin{equation}
\vspace{-5pt}
A(G,n)=100\frac{T_{\mathrm{token}}(G,n)-T_{\mathrm{shared}}(G,n)}
{T_{\mathrm{token}}(G,n)}.
\label{eq:time-advantage}
\end{equation}
Positive values indicate that the shared-token pipeline completes the run faster, while negative values favor the single-token pipeline. One value is computed for each drawing and robot count. The corresponding box plots summarize variation across drawings: boxes span the first and third quartiles, center lines denote medians, and whiskers extend to the most extreme observations within $1.5$ times the interquartile range. Observations beyond the whiskers are not displayed individually. The green markers and connecting line trace the median across robot counts, while the dashed horizontal line denotes equal completion times. These distributions characterize across-drawing variation rather than repeated-run variability.

For robot-time composition, the recorded time in each category is summed across all robots within a run and divided by the total available robot time for that run. The four resulting fractions sum to one. For each pipeline and robot count, the plotted percentage is the arithmetic mean of the corresponding per-drawing fractions, multiplied by 100, so each drawing contributes equally rather than being weighted by its total robot time.

The four robot-time categories are mutually exclusive. Drawing time consists of motion along required drawing geometry. Non-drawing movement consists of nominal transit outside the reactive avoidance protocol. Avoidance includes sidestep motion, waiting while the priority robot passes, and recovery motion back to the recorded anchor; avoidance waiting is therefore not counted as idle, and recovery motion is not counted as nominal movement. Idle time includes periods in which a robot performs none of these actions, including after it completes its assigned work or when it receives no work. Time is accounted for over the same common run interval for all robots. For a completed run, the category durations for each robot sum to the run completion time, so total available robot time is $nT_m(G,n)$.

For structural analysis, we measure the difference in within-pipeline speedup,
\vspace{-5pt}
\begin{equation}
\vspace{-5pt}
\Delta S(G,n)=S_{\mathrm{shared}}(G,n)-S_{\mathrm{token}}(G,n).
\label{eq:speedup-difference}
\end{equation}
A positive value indicates that shared-token allocation achieves a larger speedup relative to its own single-robot baseline than single-token allocation achieves relative to its baseline. Unlike $A(G,n)$, $\Delta S(G,n)$ is not a percentage reduction in completion time and does not by itself establish which pipeline finishes sooner. In particular, $\Delta S(G,1)=0$ by construction even when the two single-robot completion times differ.

For each structural attribute, drawings are ranked across the full corpus and divided into low, medium, and high levels using percentile thresholds at $1/3$ and $2/3$. Average ranks are used for tied values, so identical attribute values are not split solely to equalize group sizes. The resulting group assignment is fixed across robot counts. For each attribute level and robot count, the plotted value is the arithmetic mean of $\Delta S$ across drawings in that group.

All 385 drawings are retained in each reported structural comparison. Because tied values are kept together, the three groups need not contain exactly equal numbers of drawings. For each attribute and robot count, the sample-count-weighted mean of the low-, medium-, and high-group values reproduces the aggregate mean $\Delta S$ over the same drawing set, which is also equal to the difference between the mean shared-token and single-token speedups.

\begin{figure}[t]
\centering
\includegraphics[width=1\columnwidth]{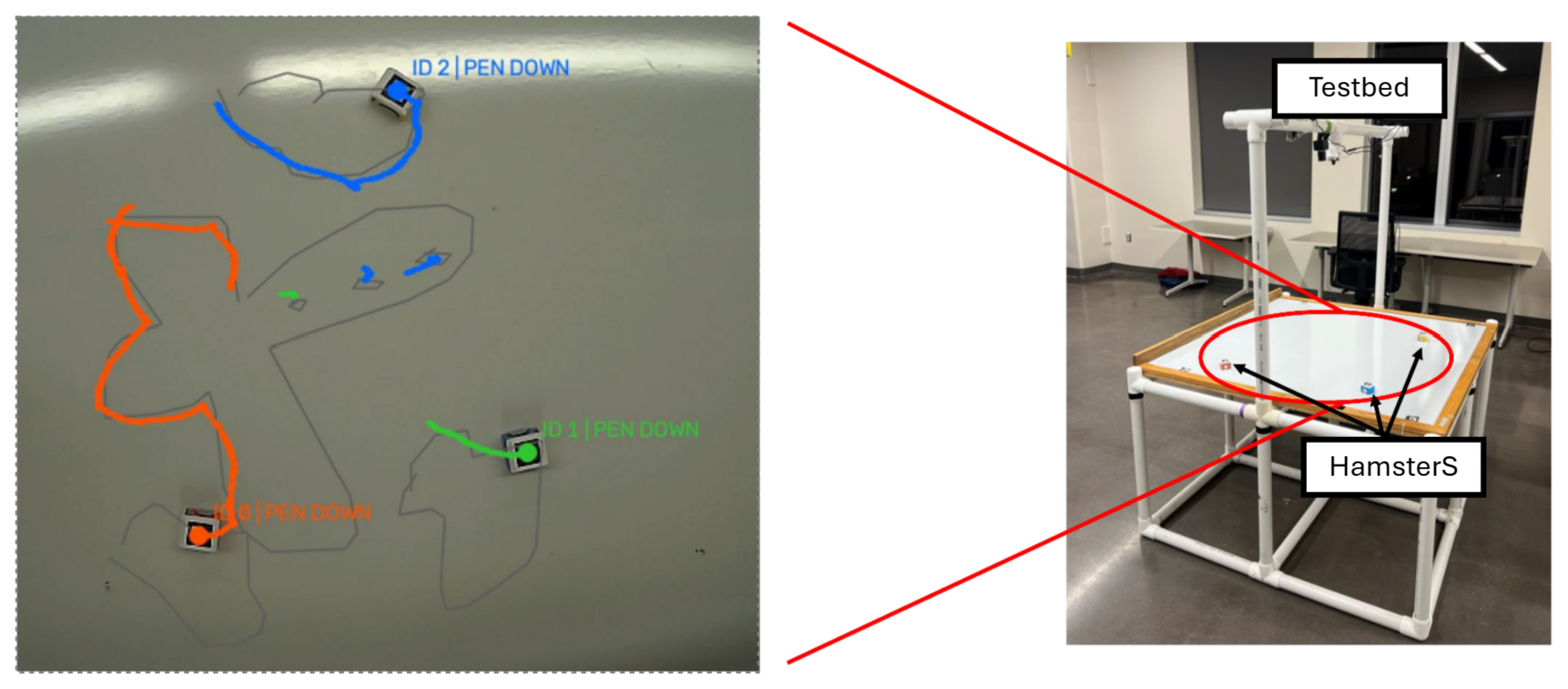}
\vspace{-10pt}
\caption{Physical multi-robot authoring setup (right) with three HamsterS robots (left) collaboratively executing the selected airplane drawing.}
\label{fig:physical_e}
\end{figure}

\subsection{Physical Experimental Setup}
\label{sec:physical-setup}
Physical validation was conducted using Robomation HamsterS robots equipped with ArUco markers and tracked by an overhead camera. The tracked robot-center trajectories were rendered as a virtual drawing overlay to visualize and verify execution of the assigned drawing paths. The same graph representation and allocation pipelines used in simulation were used to generate the routes executed by the physical robots. Robot dimensions and interaction distances were not scaled proportionally, resulting in a larger effective robot footprint than in the simulation. One representative airplane drawing was evaluated using teams of one, two, and three robots. Before each trial, the robots were placed at individual staging positions rather than the shared simulation depot. Completion time was measured from the start of execution until all robots completed their assigned work and returned to their respective staging positions. The comparison is intended to assess scaling behavior rather than exact completion-time agreement.

Figure~\ref{fig:physical_e} shows a representative three-robot trial. Speedup across robot counts and allocation methods was then compared with the corresponding simulation results for the same drawing. The physical setup and representative robot executions are also shown in the supplementary video.

\section{Results}
\label{results}

We first examine scaling with robot count, then compare paired completion times and robot-time composition between the two pipelines. Finally, we analyze how differences in within-pipeline speedup vary across structural attribute levels. This organization distinguishes the scaling question in RQ1 from the direct completion-time comparison in RQ2 and the structural associations examined in RQ3.

\subsection{Multi-Robot Scaling}
\label{results_scaling}

\begin{figure}[t]
\centering
\includegraphics[width=0.9\columnwidth]{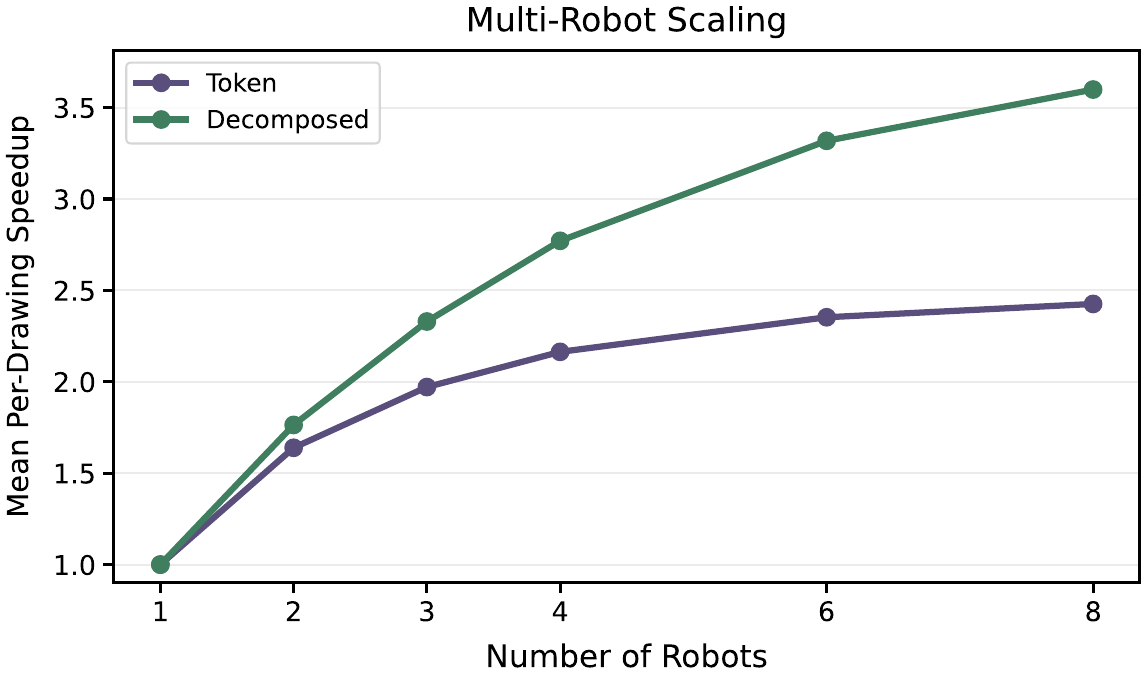}
\vspace{-5pt}
\caption{Mean per-drawing speedup relative to each pipeline's own single-robot completion time. Token and Decomposed denote single-token and shared-token allocation, respectively.}
\vspace{-2pt}
\label{fig:scaling}
\end{figure}

\begin{figure}[t]
\centering
\includegraphics[width=0.9\columnwidth]{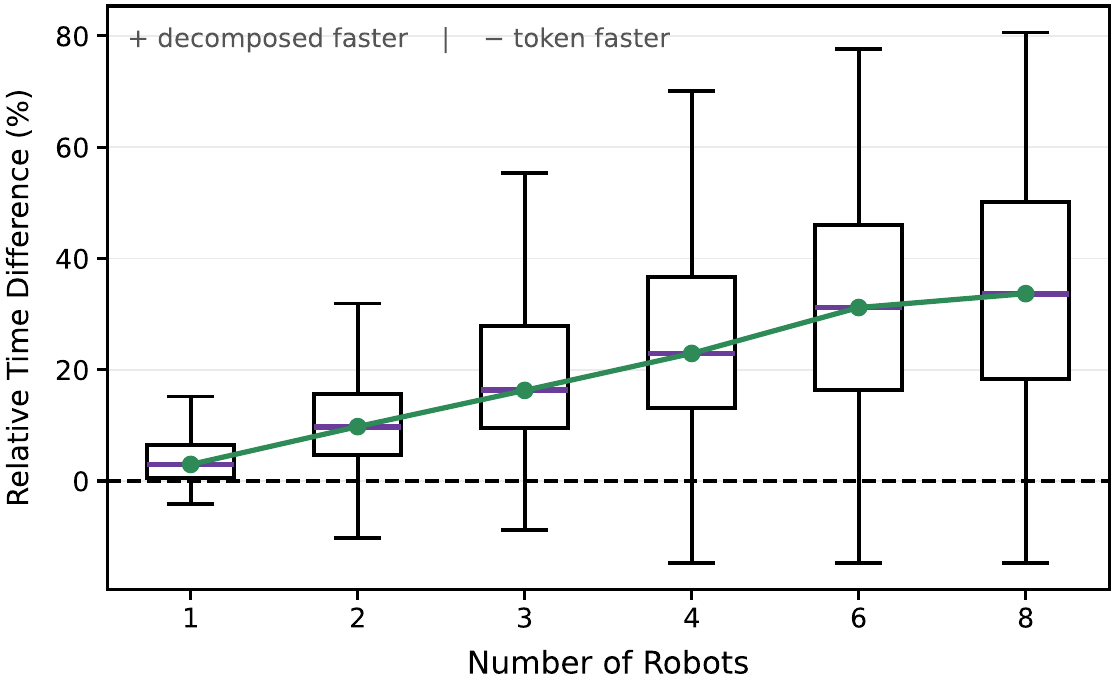}
\vspace{-5pt}
\caption{Paired relative completion-time differences $A(G,n)$ across drawings. Green markers show medians; positive values favor shared-token allocation, negative values favor single-token allocation, and the dashed line indicates equal completion times.}
\vspace{-0pt}
\label{fig:allocation_advantage}
\end{figure}

Figure~\ref{fig:scaling} shows increasing mean per-drawing speedup for both pipelines over the evaluated robot counts. Their trends diverge as team size grows: the single-token curve exhibits progressively smaller gains, whereas the shared-token curve continues to increase more strongly. At eight robots, mean speedup reaches 2.426 for the single-token pipeline and 3.599 for the shared-token pipeline. The separation is greatest at the larger team sizes, indicating different scaling behavior relative to the pipelines' respective single-robot baselines.

These trends are consistent with the different concurrency restrictions of the allocation strategies. Under single-token allocation, each connected component can be assigned to only one robot, so component count and unequal component workloads can limit simultaneous drawing. Shared-token allocation permits robots to share work within a component and therefore exposes additional opportunities for parallel execution. The observed curves describe these pipelines over the tested team sizes; they do not establish a universal scaling limit for component-level allocation.

\subsection{Allocation Strategy}
\label{results_allocation}

Figure~\ref{fig:allocation_advantage} compares the pipelines directly through the paired percentage difference $A(G,n)$. The median becomes more positive as robot count increases, while the distribution across drawings becomes more dispersed. At eight robots, the median relative completion-time difference is 33.67\%, with first and third quartiles of 18.32\% and 50.13\%, respectively. Shared-token allocation completes 372 of the 385 drawings faster, while single-token allocation completes 13 faster, with no ties. Negative values therefore remain, indicating drawings for which the single-token pipeline finishes sooner. Thus, the aggregate advantage of the shared-token pipeline does not imply a uniform completion-time improvement across the corpus.

\begin{figure}[t]
\centering
\includegraphics[width=0.9\columnwidth]{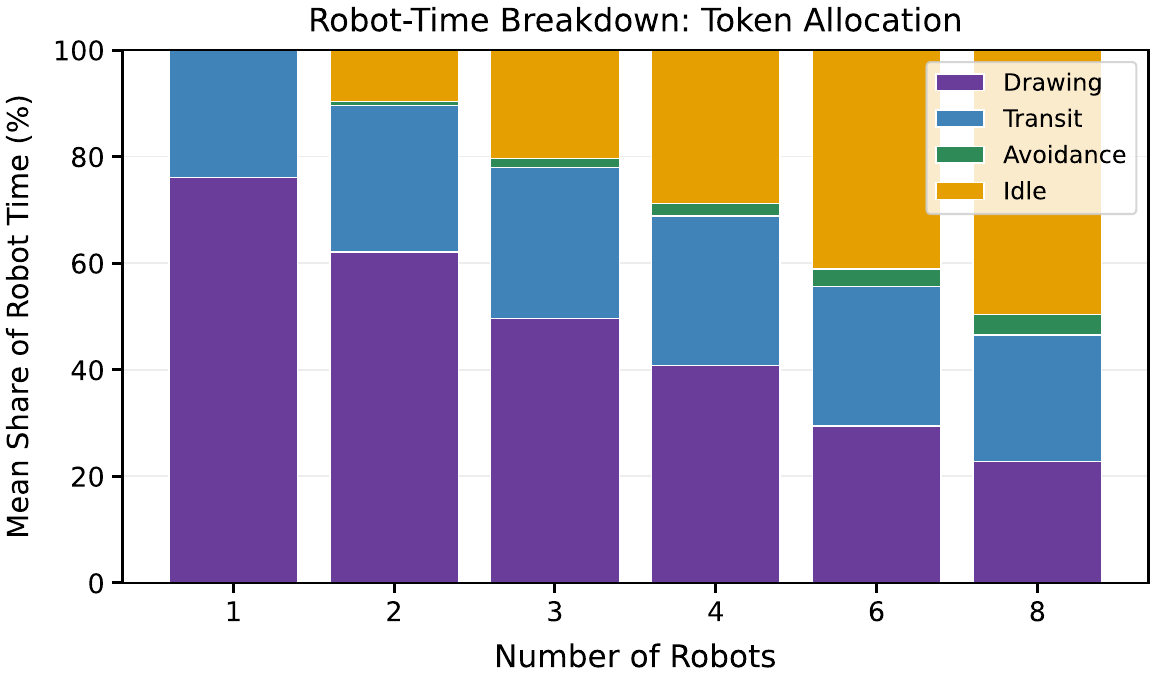}
\vspace{-5pt}
\caption{Robot-time composition under single-token allocation. Each segment is the mean of the corresponding per-drawing robot-time fractions, expressed as a percentage. Each stacked bar sums to 100\%.}
\vspace{-0pt}
\label{fig:token_time}
\end{figure}

\begin{figure}[t]
\centering
\vspace{-0pt}
\includegraphics[width=0.9\columnwidth]{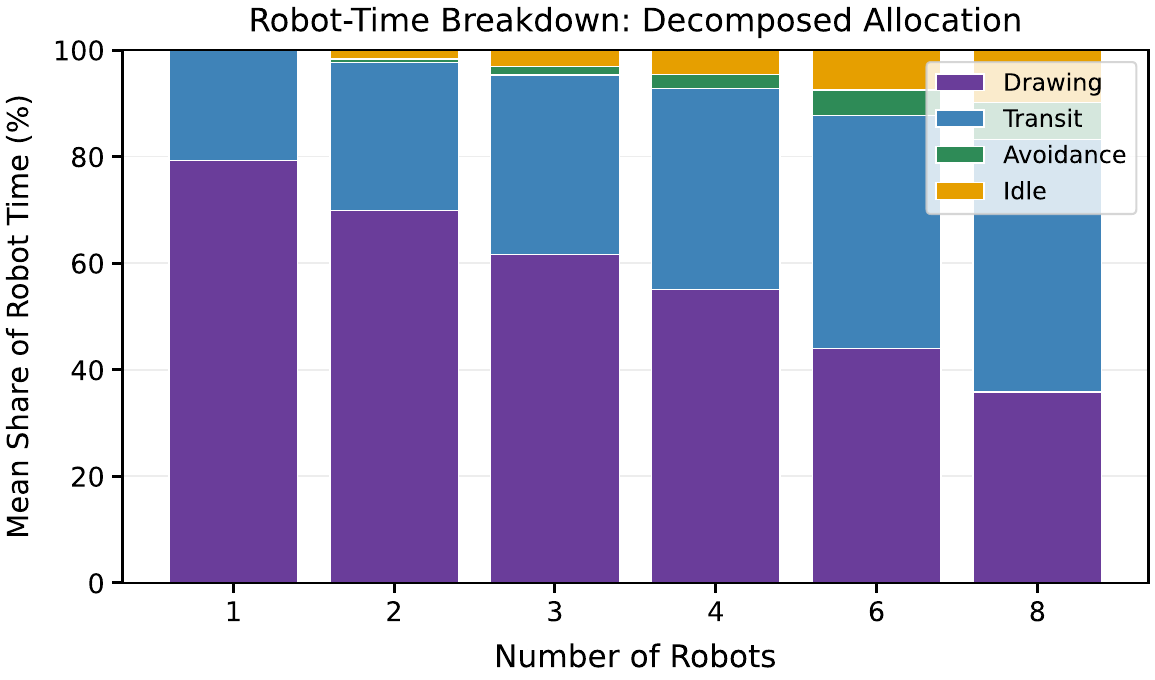}
\vspace{-5pt}
\caption{Robot-time composition under shared-token (Decomposed) allocation, using the same per-drawing normalization and averaging as Fig.~\ref{fig:token_time}.}
\vspace{-0pt}
\label{fig:decomposed_time}
\end{figure}

The robot-time compositions in Figs.~\ref{fig:token_time} and~\ref{fig:decomposed_time} show different patterns of resource utilization. Under single-token allocation, the mean idle-time share grows substantially with robot count. At eight robots, the mean per-drawing idle-time share is 49.65\% under single-token allocation and 9.80\% under shared-token allocation. At larger team sizes, shared-token allocation exhibits a smaller idle-time share and larger shares devoted to drawing, non-drawing movement, and avoidance. The lower idle fraction is consistent with greater use of the available robots, although it does not distinguish component indivisibility from workload imbalance or early completion of individual routes. Because the bars show normalized shares, larger movement and avoidance fractions do not establish larger absolute durations. Absolute robot-time measurements are needed to quantify those execution costs.

\subsection{Structural Characteristics}
\label{results_structure}

\begin{figure}[t]
\centering
\includegraphics[width=1\columnwidth]{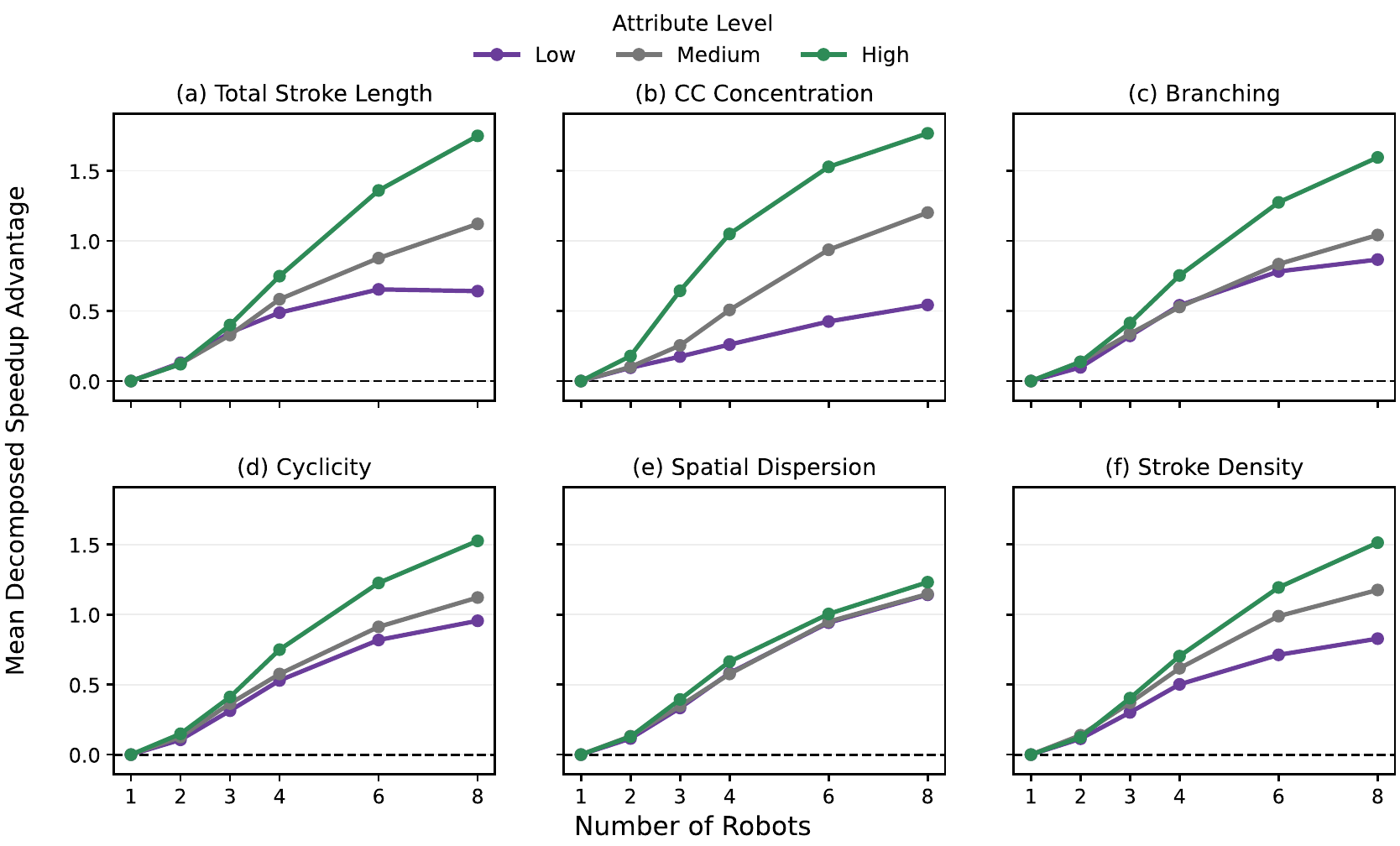}
\vspace{-15pt}
\caption{Mean within-pipeline speedup difference, $\Delta S(G,n)$, across low, medium, and high structural attribute groups.}
\vspace{-5pt}
\label{fig:all_attributes}
\end{figure}

Figure~\ref{fig:all_attributes} examines associations between task structure and the difference in within-pipeline speedup, $\Delta S(G,n)$. For each structural characteristic, drawings are divided into low, medium, and high levels according to their position within the full drawing corpus. These groups partition the same 385 drawings used in the aggregate scaling analysis and remain fixed across robot counts. Other structural characteristics can covary, so the resulting comparisons are descriptive rather than estimates of independent attribute effects.

The high-stroke-length group exhibits a larger mean $\Delta S(G,n)$ than the low-stroke-length group, particularly at larger team sizes (Fig.~\ref{fig:all_attributes}(a)). A similar and slightly stronger separation appears for component concentration $C_L$ (Fig.~\ref{fig:all_attributes}(b)). The concentration trend is consistent with the component-indivisibility restriction: when a large fraction of the required stroke length lies within one component, single-token allocation cannot distribute that component's drawing work among robots, whereas shared-token allocation can.

The high-branching, high-cyclicity, and high-stroke-density groups also exhibit larger mean $\Delta S(G,n)$ values than their corresponding low groups as robot count increases (Fig.~\ref{fig:all_attributes}(c), (d), and (f)). In contrast, the spatial-dispersion curves remain closely grouped throughout the evaluated team sizes (Fig.~\ref{fig:all_attributes}(e)), indicating substantially weaker separation based on spatial dispersion alone.

At eight robots, the high-minus-low mean $\Delta S$ contrast is largest for component concentration (1.225) and total stroke length (1.108), while spatial dispersion produces a contrast of only 0.090. Branching, cyclicity, and stroke density fall between these extremes. These results reinforce the visual pattern that workload magnitude and concentration provide stronger descriptive separation of decomposition benefit than spatial dispersion.

Across several descriptors, differences between group means become more pronounced at larger team sizes. The equality of all groups at one robot follows from the normalization in~\eqref{eq:speedup} and is not evidence that task structure has no effect on single-robot execution time. At the remaining robot counts, the panels show variation in scaling differences across structural attribute levels. These patterns motivate considering workload distribution and graph topology alongside component count when comparing allocation pipelines.

\subsection{Physical Validation}
\label{real_world}
For the selected airplane drawing, the physical experiments showed the same qualitative scaling trend as the corresponding simulation runs as summarized in Table~\ref{tab:physical_sim_speedup}. Token allocation improved from one to two robots but showed little additional benefit at a higher count of  three robots. Decomposed allocation continued to scale with increasing team size. The denser physical setting provides additional context for the reduced speedup at only three robots.

\begin{table}[h]
\centering
\vspace{-0pt}
\caption{Comparison of simulated and physical speedup}
\label{tab:physical_sim_speedup}
\begin{tabular}{c c c c c}
\hline
\textbf{Method} & \textbf{Robots} & \textbf{Simulation} & \textbf{Physical} & \textbf{Difference} \\
\hline
Token & 2 & 1.71$\times$ & 1.63$\times$ & $-4.8\%$ \\
Token & 3 & 1.71$\times$ & 1.59$\times$ & $-7.3\%$ \\
Decomposed & 2 & 1.76$\times$ & 1.69$\times$ & $-3.7\%$ \\
Decomposed & 3 & 2.31$\times$ & 2.10$\times$ & $-9.1\%$ \\
\hline
\end{tabular}
\vspace{-5pt}
\end{table}

\section{Conclusion}
This paper examined task-dependent scaling in multi-robot authoring through a simulation study of 385 stroke graphs and six robot-team sizes. With respect to scaling, both allocation-routing pipelines benefited from additional robots, but the single-token pipeline exhibited increasingly diminishing gains while the shared-token pipeline continued to scale more strongly at larger team sizes. With respect to direct completion time, shared-token allocation was faster for most drawings at larger team sizes, although the advantage was not universal. For task structure, total stroke length and component concentration produced the strongest separation in scaling differences, while branching, cyclicity, and stroke density showed intermediate separation, and spatial dispersion produced the weakest effect. These findings characterize the evaluated allocation-routing pipelines rather than isolate the causal effect of decomposition granularity. They highlight the importance of considering within-component workload and graph structure when selecting an allocation pipeline and robot-team size. The simulation results include runs with geometric robot overlap and should therefore not be interpreted as collision-free completion-time measurements.


\section*{Acknowledgment}
OpenAI's ChatGPT was used to assist with development and debugging of experimental software and data processing. All AI-assisted content was reviewed and validated by the authors.

\bibliographystyle{IEEEtran}
\bibliography{references}
\end{document}